\documentclass[sigconf,nonacm]{acmart}

\usepackage[dvipsnames,table]{xcolor}
\usepackage{amsmath}
\usepackage{booktabs,longtable,tabularx,array,calc,multirow,arydshln}
\usepackage{nicefrac,microtype,graphicx,etoolbox,needspace}
\usepackage{float}
\usepackage{placeins}
\usepackage{algorithm}
\usepackage{algpseudocode}
\floatname{algorithm}{Algorithm}
\algnewcommand\algorithmicinput{\textbf{Input}}
\algnewcommand\Input{\item[\algorithmicinput]}
\algnewcommand\algorithmicoutput{\textbf{Output}}
\algnewcommand\Output{\item[\algorithmicoutput]}
\algtext*{EndFor}
\algtext*{EndIf}
\algtext*{EndWhile}
\usepackage{bookmark}
\IfFileExists{xurl.sty}{\usepackage{xurl}}{}

\IfFileExists{footnotehyper.sty}{\usepackage{footnotehyper}}{\usepackage{footnote}}

\providecommand{\tightlist}{%
  \setlength{\itemsep}{0pt}\setlength{\parskip}{0pt}}
\hypersetup{
  pdftitle={Contrastive Branch Policy Optimization},
  pdfauthor={Ying Wang, Changlin Qiu, Bang Lin, Linbo Jin, Wen Jiang, Zhe Sun, Jingli Yang}
}

\newcommand{\cbpomainfont}{\small}

\newcommand{\cbpoauthorblock}{%
  \vbox{%
    \hsize=0.98\textwidth
    \centering
    {\large\bfseries
      \makebox[\hsize][c]{%
        Ying Wang\textsuperscript{1,2}\hspace{1em}
        Changlin Qiu\textsuperscript{1,*}\hspace{1em}
        Bang Lin\textsuperscript{1}\hspace{1em}
        Linbo Jin\textsuperscript{1}\hspace{1em}
        Wen Jiang\textsuperscript{1}\hspace{1em}
        Zhe Sun\textsuperscript{1}\hspace{1em}
        Jingli Yang\textsuperscript{2}}%
      \par
    }
    \vspace{0.65em}
    {\normalsize\normalfont
      \makebox[\hsize][c]{%
        \textsuperscript{1}Alibaba Group, Hangzhou, China\hspace{1.2em}
        \textsuperscript{2}Harbin Institute of Technology, Harbin, China}%
      \par
    }
  }
}
\title{Contrastive Branch Policy Optimization}
\author{Ying Wang}
\affiliation{%
  \institution{Alibaba Group}
  \city{Hangzhou}
  \country{China}
}
\affiliation{%
  \institution{Harbin Institute of Technology}
  \city{Harbin}
  \country{China}
}
\email{24s005080@stu.hit.edu.cn}

\author{Changlin Qiu}
\authornote{Corresponding author: Changlin Qiu (\href{mailto:qiuchanglin.qcl@taobao.com}{qiuchanglin.qcl@taobao.com}).}
\affiliation{%
  \institution{Alibaba Group}
  \city{Hangzhou}
  \country{China}
}
\email{qiuchanglin.qcl@taobao.com}

\author{Bang Lin}
\affiliation{%
  \institution{Alibaba Group}
  \city{Hangzhou}
  \country{China}
}
\email{linbang.lb@taobao.com}

\author{Linbo Jin}
\affiliation{%
  \institution{Alibaba Group}
  \city{Hangzhou}
  \country{China}
}
\email{yuyi.jlb@alibaba-inc.com}

\author{Wen Jiang}
\affiliation{%
  \institution{Alibaba Group}
  \city{Hangzhou}
  \country{China}
}
\email{wen.jiangw@alibaba-inc.com}

\author{Zhe Sun}
\affiliation{%
  \institution{Alibaba Group}
  \city{Hangzhou}
  \country{China}
}
\email{sz481403@alibaba-inc.com}

\author{Jingli Yang}
\affiliation{%
  \institution{Harbin Institute of Technology}
  \city{Harbin}
  \country{China}
}
\email{jinglidg@hit.edu.cn}
\renewcommand{\shortauthors}{Ying Wang, Changlin Qiu, Bang Lin, Linbo Jin, Wen Jiang, Zhe Sun, and Jingli Yang}

\makeatletter
\gdef\addresses{\@author{\cbpoauthorblock}}
\renewcommand{\@mktitle@iii}{%
  \hsize=\textwidth
  \setbox\mktitle@bx=\vbox{%
    \vspace*{-0.65em}%
    \@titlefont\centering
    \@ACM@title@width=\hsize
    \parbox[t]{\@ACM@title@width}{%
      \centering\@titlefont\@title\@translatedtitle
      \ifx\@subtitle\@empty\else
        \par\noindent{\@subtitlefont\@subtitle\@translatedsubtitle}%
      \fi
    }%
    \par\vspace{1.6em}%
  }%
}
\makeatother

\begin{document}

\begin{abstract}
Reinforcement learning with verifiable rewards (RLVR) enables language models to learn multi-turn interaction with external tools, yet its sparse outcome rewards provide no signal for identifying which intermediate decisions are responsible for success. Branch sampling induces local comparisons among alternative continuations, but existing methods tend to conflate two distinct problems: allocating a fixed rollout budget and translating branch outcomes into token-level credit. We introduce Contrastive Branch Policy Optimization (CBPO), which disentangles these two problems and assigns a dedicated mechanism to each. Generation entropy screens candidate branch positions across the entire response, while path-level and node-level decay distribute a fixed budget across trajectories and positions to prevent exploration from collapsing onto a few paths or adjacent tokens. A parent trajectory together with the branches that share an identical token prefix forms an exact-prefix group, and the reward variation within this controlled group defines the Contrastive Branch Value (CBV), an outcome-based estimate of local decision sensitivity that rescales continuation advantages without altering their sign. When multiple nodes are selected along the same trajectory, CBPO partitions it into non-overlapping credit segments, thereby avoiding duplicated gradients on shared tokens. Requiring only outcome rewards and no process-level annotation, CBPO provides a practical solution for fine-grained credit assignment in tool-integrated agent training. Extensive experiments on ten benchmarks---five for mathematical reasoning and five for knowledge-intensive search---show that CBPO consistently outperforms state-of-the-art policy-optimization and branch-based methods, attaining the highest macro-average accuracy in both domains and across two model scales.
\end{abstract}

\keywords{reinforcement learning, large language models, tool-integrated reasoning, credit assignment}
\begin{CCSXML}
<ccs2012>
<concept>
<concept_id>10010147.10010257.10010258.10010261</concept_id>
<concept_desc>Computing methodologies~Reinforcement learning</concept_desc>
<concept_significance>500</concept_significance>
</concept>
<concept>
<concept_id>10010147.10010178.10010219.10010221</concept_id>
<concept_desc>Computing methodologies~Intelligent agents</concept_desc>
<concept_significance>300</concept_significance>
</concept>
</ccs2012>
\end{CCSXML}
\ccsdesc[500]{Computing methodologies~Reinforcement learning}
\ccsdesc[300]{Computing methodologies~Intelligent agents}

\maketitle

% The ACM class controls typography and column spacing.

\section{Introduction}\label{sec:introduction}

Chain-of-thought prompting \cite{wei2022chain} and reinforcement learning with verifiable rewards (RLVR) \cite{schulman2017ppo,shao2024deepseekmath} have substantially improved the reasoning capabilities of large language models (LLMs). Many tasks, however, require current information, exact computation, or feedback from an external environment. Addressing such tasks entails multi-turn interaction with search engines, code interpreters, or both \cite{yao2023react,gou2024tora}. Recent systems apply reinforcement learning to these interactions \cite{feng2026retool,li2025torl}, including settings that require coordinated use of multiple tools \cite{dong2025toolstar}.

The resulting training signal nevertheless remains coarse. Most RLVR and agentic reinforcement-learning methods assign a single terminal reward to every generated token in a trajectory. Such uniform credit cannot distinguish a decisive tool request or correction from text that merely precedes a successful answer. Process supervision \cite{lightman2024verify} and step-wise preference optimization \cite{lai2024stepdpo} provide finer signals but require intermediate labels or preferences. Tree-based sampling instead compares continuations from a shared history using only outcome rewards \cite{hou2025treerl}. This approach, however, still requires a principled policy for locating branches under a limited budget and mapping their outcomes to local policy updates.

Existing methods address these requirements only partially. GIGPO and Tree-GRPO construct local advantages from shared or tree-structured histories \cite{feng2025gigpo,ji2026treesearch}, whereas ARPO uses generation entropy to select branch points after tool feedback \cite{dong2025arpo}. Entropy provides an inexpensive proxy for token-level uncertainty \cite{lin2024critical,cheng2025exploration,wang2025entropyminority}, but does not measure outcome sensitivity. Allocation based solely on entropy can concentrate branches at adjacent positions, overlook consequential decisions outside tool boundaries, or emphasize variation that leaves the final answer unchanged.

Contrastive Branch Policy Optimization (CBPO) separates candidate discovery from credit assignment. CBPO scans local entropy throughout the response and applies path-level and node-level decay to distribute a fixed branch budget. A parent trajectory and branches sharing an identical token prefix form an exact-prefix group. Reward variation within this controlled group defines Contrastive Branch Value (CBV), an outcome-based estimate of local decision sensitivity. A bounded form of CBV changes the magnitude, but not the sign, of continuation advantages. Copied prefixes receive no branch gradient, and multiple branch nodes partition a parent trajectory into non-overlapping credit segments. The two signals therefore serve distinct functions: entropy identifies alternative continuations, whereas observed outcome variation estimates their task relevance.

The evaluation spans two model scales and includes five tool-augmented mathematical benchmarks and five knowledge-intensive search benchmarks. CBPO consistently outperforms state-of-the-art policy-optimization and branch-based methods, attaining the highest macro average in both domains.

The main contributions are as follows:

\begin{itemize}
\tightlist
\item
  Branch selection is formulated as budgeted exploration over the entire response. Fixed-interval entropy windows expose candidates beyond tool boundaries, while path-level and node-level decay prevent allocation from collapsing onto a few trajectories or adjacent positions.
\item
  CBV estimates local decision sensitivity from reward variation within exact-prefix groups. Standardized and bounded CBV modulates continuation advantages without changing their signs. Prefix masking and non-overlapping segmentation prevent repeated credit on shared tokens.
\item
  Experiments across ten benchmarks and two model scales show that CBPO consistently outperforms strong policy-optimization and branch-based baselines. Cross-scale ablations further attribute these gains to balanced branch allocation and outcome-contrastive credit assignment.
\end{itemize}

\section{Related Work}\label{sec:related-work}

\subsection{Fine-Grained Credit Assignment}\label{sec:fine-grained-credit}

PPO optimizes a policy from trajectory-level returns \cite{schulman2017ppo}. GRPO-based reasoning systems likewise derive one advantage from each completed response \cite{shao2024deepseekmath,guo2025deepseekr1}. Subsequent variants improve optimization stability or scaling \cite{yu2025dapo,hu2025reinforcepp}, while GSPO moves the optimization unit from tokens to sequences \cite{zheng2025gspo}. These methods train directly from outcome rewards but do not distinguish decisive intermediate choices from weakly related transitional text. Problem decomposition, computational correction, tool selection, and answer verification can contribute differently to a multi-step solution. Assigning a single advantage may therefore reinforce effective decisions and incidental behavior together.

Process rewards \cite{lightman2024verify} and step-wise preference optimization \cite{lai2024stepdpo} supervise intermediate reasoning more directly. Preference objectives such as DPO \cite{rafailov2023dpo} and token-entropy methods \cite{he2026polarityentropy} offer alternative local signals. Process labels and learned value estimates can provide dense feedback but entail annotation costs or estimation error. Entropy requires neither annotation nor learned value estimation, yet measures predictive uncertainty rather than a position's empirical association with the final reward.

Tree-based methods reuse prefixes and sample several continuations from intermediate states, enabling local comparisons at controlled cost. Online variants derive advantages from current-policy descendants, within-tree contrasts, or recurring histories \cite{hou2025treerl,ji2026treesearch}. Entropy can identify uncertain candidates at low computational cost \cite{lin2024critical,cheng2025exploration}. Moreover, a small fraction of high-entropy tokens can dominate effective updates \cite{wang2025entropyminority}. EAPO combines reward polarity with token entropy. AEPO balances entropy during rollout and optimization while limiting repeated branches at consecutive high-entropy positions \cite{he2026polarityentropy,dong2025aepo}. These approaches use uncertainty to guide optimization, but divergent token distributions need not yield divergent outcomes. CBPO instead restricts entropy to candidate screening and estimates local credit from outcome contrasts among continuations sharing an identical prefix.

\subsection{Tool-Integrated Reasoning}\label{sec:tool-integrated-reasoning}

External tools allow language models to access current information, perform exact computation, and act on an environment \cite{yao2023react,lin2025tir}. Tool-integrated reasoning extends beyond the binary decision to invoke a tool. A model must select a tool, construct the request, determine when to invoke it, verify the response, integrate the result, and terminate appropriately. Search provides open-world evidence, whereas code interpreters support calculation and programmatic verification. Coordinating both requires repeated transitions between internal reasoning and external actions as new observations arrive.

Work in this area has progressed from prompted reasoning and acting \cite{yao2023react,gou2024tora} to reinforcement learning for search \cite{jin2025searchr1,sun2025zerosearch}, code interpreters \cite{feng2026retool,li2025torl}, and multi-tool coordination \cite{dong2025toolstar}. Recent WSDM studies train autonomous programmatic agents and optimize tool selection with reinforcement learning \cite{jiang2026tablemind,zhang2026toolcure}. Unlike prompting or supervised imitation, reinforcement learning can optimize the complete interaction from environmental feedback and terminal outcomes. In long trajectories, however, success may depend on pre-call reasoning, request construction, feedback interpretation, or post-call synthesis. Tool-return boundaries therefore capture only a subset of potentially decisive positions.

Knowledge-intensive tasks also depend on retrieval quality and timing. Adaptive retrieval can extend beyond an initial ranking when the first-stage candidate pool has insufficient recall \cite{rathee2025quam}. In LLM-based fact-checking, adaptive evidence retrieval allows a model to determine when internal knowledge is inadequate or conflicts with external evidence \cite{zhang2026knowfc}. These WSDM studies concern retrieval and verification rather than token-level credit assignment, but reveal the same structural challenge. An agent must determine where additional computation or evidence can alter the final outcome.

Long-horizon tool learning often improves planning or attribution through reward shaping, process supervision, or step-level optimization. Intermediate supervision is costly, while predefined states and step boundaries constrain the granularity of credit assignment. GIGPO derives step-level advantages from shared histories. ARPO branches after tool feedback and assigns the resulting advantage to the suffix \cite{feng2025gigpo,dong2025arpo}. The former depends on repeated history states, and the latter treats tool returns as fixed branch boundaries. CBPO searches the full response for branch candidates, then attributes local credit by comparing outcomes under an identical prefix.

\section{Preliminaries}\label{sec:preliminaries}

\textbf{Tool-integrated rollout.}
Given a problem \(x\) drawn from dataset \(\mathcal{D}\) and a tool set \(\mathcal{T}\), policy \(\pi_\theta\) generates an interaction trajectory \(\tau\) that interleaves model tokens, tool requests, and environmental observations. Let \(z=(z_1,\ldots,z_{|z|})\) collect the model-generated tokens of \(\tau\), and let \(o_{<t}\) denote the tool observations available before position \(t\). Trajectory generation then factorizes autoregressively over model tokens only:

\begin{equation}
P_\theta(\tau\mid x;\mathcal{T})
=\prod_{t=1}^{|z|}
\pi_\theta\!\left(z_t\mid x,z_{<t},o_{<t};\mathcal{T}\right).
\label{eq:rollout-factorization}
\end{equation}

Observations are inserted by the environment upon each tool call; they condition subsequent generation but contribute neither likelihood terms nor policy gradients. Upon termination, a verifier assigns a bounded outcome reward \(R(\tau)\). Following agentic reinforcement learning formulations \cite{dong2025arpo,dong2025aepo}, training maximizes the KL-regularized expected reward:

\begin{equation}
\max_{\pi_\theta}\;
\mathbb{E}_{x\sim\mathcal{D},\,\tau\sim\pi_\theta(\cdot\mid x;\mathcal{T})}
\bigl[R(\tau)\bigr]
-\beta\,D_{\mathrm{KL}}\!\left(\pi_\theta\,\Vert\,\pi_{\mathrm{ref}}\right),
\label{eq:agentic-objective}
\end{equation}

where \(\pi_{\mathrm{ref}}\) is a frozen reference policy and \(\beta\) controls the regularization strength. Shared-history, suffix-based, and tree-search agent RL methods adopt this trajectory-level reward setting \cite{feng2025gigpo,ji2026treesearch,dong2025arpo}.

\textbf{Group-relative optimization.}
For a given problem, Group Relative Policy Optimization (GRPO) samples \(G\) trajectories and estimates the advantage of trajectory \(k\) as \cite{shao2024deepseekmath}:

\begin{equation}
A_k=
\frac{R(\tau_k)-\mu_R}
{\sigma_R+\epsilon},
\qquad
\mu_R=\frac{1}{G}\sum_{j=1}^{G}R(\tau_j),
\label{eq:grpo-advantage}
\end{equation}

Here, \(\mu_R\) and \(\sigma_R\) are the within-group reward mean and standard deviation, and \(\epsilon\) provides numerical stability. Standard GRPO assigns \(A_k\) to every model-generated token in trajectory \(k\), so the update magnitude is uniform along the trajectory regardless of which intermediate decisions determine the outcome.

\textbf{Generation uncertainty.}
At generation position \(t\), let the context be \(h_t=(x,z_{<t})\) and the next-token distribution be \(p_t=\pi_\theta(\cdot\mid h_t)\). Its Shannon entropy is

\begin{equation}
H_t=-\sum_{v\in\mathcal{V}}p_{t,v}\log p_{t,v},
\label{eq:token-entropy}
\end{equation}

where \(\mathcal{V}\) denotes the vocabulary. Larger \(H_t\) indicates greater uncertainty over the next token. This quantity reflects the dispersion of the generation distribution at position \(t\) rather than the confidence of any particular sampled token. Prior work uses token entropy to identify critical decisions \cite{lin2024critical} and characterize exploration in language-model reasoning \cite{cheng2025exploration,wang2025entropyminority}.

\section{Contrastive Branch Policy Optimization}\label{sec:method}

\subsection{Overview}\label{sec:method-overview}

\begin{figure*}[!t]
\centering
\includegraphics[width=\textwidth]{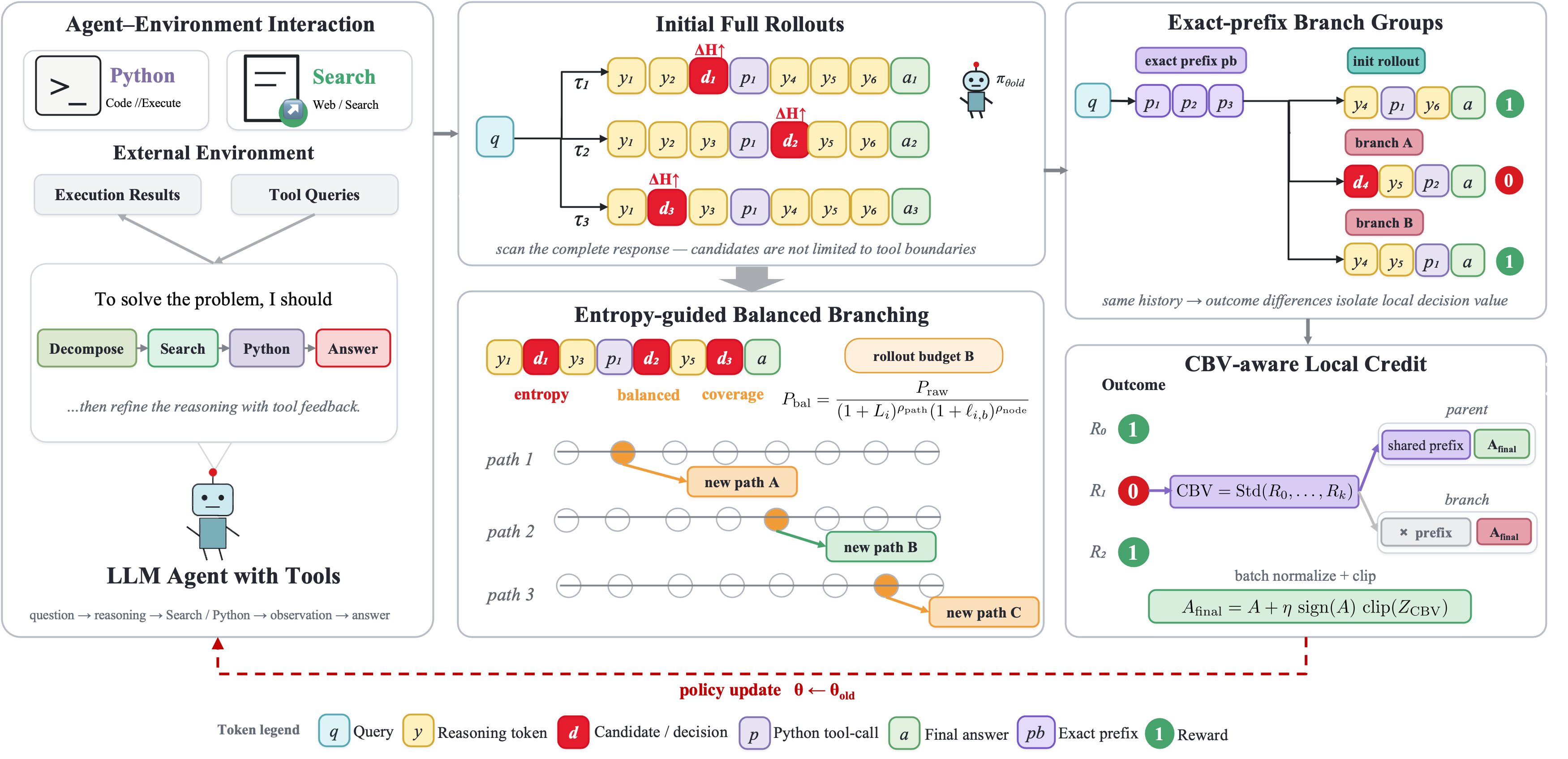}
\caption{CBPO training framework. The model first performs interactive reasoning with Python and Search tools and samples complete initial trajectories. Candidate nodes are identified across the full response, and path-level and node-level decay allocate the branch budget. Outcome rewards estimate CBV within exact-prefix groups. Shared-prefix deduplication and non-overlapping hierarchical segmentation then construct token-level advantages for the policy update.}
\Description{End-to-end CBPO pipeline from tool-integrated trajectory sampling through balanced branching, CBV estimation, and policy optimization.}
\label{fig:cbpo-overview}
\end{figure*}

CBPO addresses two coupled limitations of branch-based RL: concentrated exploration and imprecise local credit. The method uses generation uncertainty for candidate selection and observed outcome variation to modulate update magnitude. Figure~\ref{fig:cbpo-overview} summarizes the resulting training framework. Starting from complete interaction trajectories, CBPO scans each response for uncertain positions and allocates a fixed branch budget through path-level and node-level decay. Rewards are then compared among continuations sharing an exact prefix to construct CBV-aware local credit. Shared-prefix deduplication and non-overlapping segmentation convert this credit into token-level advantages. Section~\ref{sec:balanced-branching} constructs balanced exact-prefix groups, and Section~\ref{sec:cbv-optimization} converts their outcomes into policy updates.

The design maintains three invariants. First, every problem receives the same total rollout budget because unused branch slots are reallocated to independent complete trajectories. Second, each local comparison conditions on an identical token history, so reward variation arises only from resampled continuations. Third, copied prefixes are excluded from branch losses, and parent trajectories are partitioned into non-overlapping intervals when several nodes are selected. In tool-integrated trajectories, environmental observations condition later actions but are neither scanned nor assigned policy gradients, since they are not model-generated. These constraints separate allocation from attribution and prevent additional branches from duplicating shared-token gradients.

\subsection{Entropy-Guided Balanced Branch Exploration}\label{sec:balanced-branching}

\begin{figure}[!t]
\centering
\includegraphics[width=0.96\linewidth]{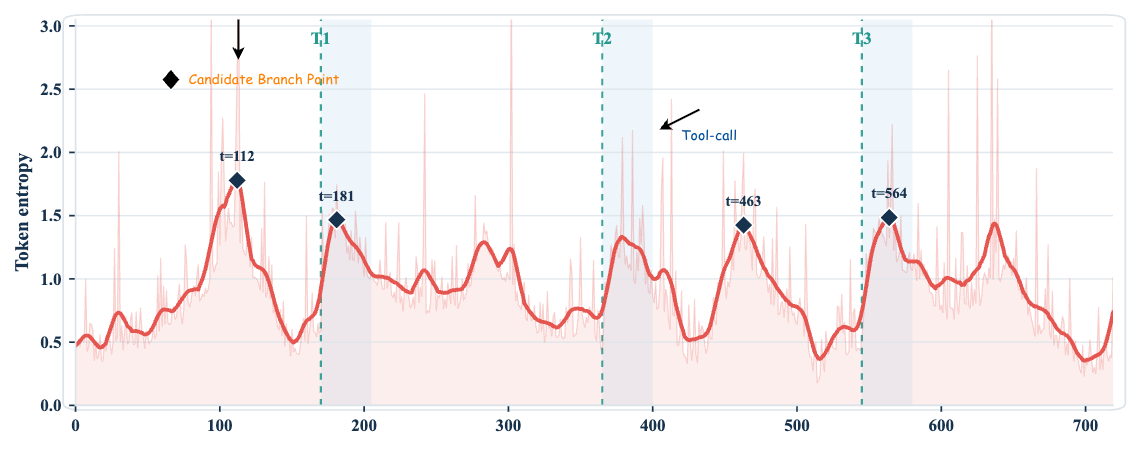}
\caption{Full-response candidate discovery. Fixed-interval entropy scans identify high-entropy decisions beyond predefined tool-call boundaries.}
\Description{Line plot of token entropy across a full response, with fixed-interval windows, tool-call positions, and selected candidate branch points.}
\label{fig:candidate-discovery}
\end{figure}

Exhaustive branching at every token is computationally infeasible, whereas branching only at tool-return boundaries assumes that consequential decisions coincide with predefined interaction events. Entropy provides an inexpensive candidate signal, but an unregularized ranking can allocate most of a fixed budget to a few trajectories or neighboring positions. This stage therefore combines full-response candidate discovery with explicit coverage control. The procedure first samples \(N_0\) complete trajectories from the current policy. Figure~\ref{fig:candidate-discovery} illustrates how fixed-interval scanning identifies high-entropy decisions beyond tool-call boundaries.

\textbf{Full-trajectory candidate identification.}
For trajectory \(i\), CBPO places candidate boundaries throughout the response at intervals of \(d_{\min}\):

\begin{equation}
\mathcal{B}_i
=
\{b_{i,q}=q d_{\min}\}_{q=1}^{Q_i},
\label{eq:candidate-boundaries}
\end{equation}

where \(Q_i\) is the number of valid candidates. Let \(\mathcal{W}_{i,b}\) denote the window of \(w\) model-generated tokens beginning at boundary \(b\in\mathcal{B}_i\). Its normalized entropy is

\begin{equation}
H_{i,b}
=
-\frac{1}{|\mathcal{W}_{i,b}|\log|\mathcal{V}|}
\sum_{t\in\mathcal{W}_{i,b}}
\sum_{v\in\mathcal{V}_{K}^{i,t}}
p_{i,t,v}\log p_{i,t,v},
\label{eq:window-entropy}
\end{equation}

where \(\mathcal{V}_{K}^{i,t}\) is the top-\(K\) candidate set at token \(t\), and \(p_{i,t,v}\) is the probability of candidate \(v\). The retained probability mass is not renormalized. Thus, \(H_{i,b}\) is a truncated entropy proxy used only to rank candidate locations, not the entropy of a top-\(K\) distribution.

The first window of each trajectory provides a path-specific reference:

\begin{equation}
H_i^{\mathrm{root}}=H_{i,0}.
\label{eq:root-entropy}
\end{equation}

Local uncertainty at boundary \(b\) is measured by the entropy increase relative to this reference:

\begin{equation}
\Delta H_{i,b}=H_{i,b}-H_i^{\mathrm{root}}.
\label{eq:relative-entropy-increase}
\end{equation}

The corresponding raw priority is

\begin{equation}
P_{i,b}^{\mathrm{raw}}
=
\operatorname{clip}
\left(
\alpha+\gamma\Delta H_{i,b},0,1
\right),
\label{eq:raw-priority}
\end{equation}

where \(\alpha\) sets the base priority, \(\gamma\) scales the entropy increase, and \(\operatorname{clip}\) restricts the score to \([0,1]\).

\textbf{Path-node balanced budgeting.}
An entropy ranking alone can allocate most branches to a few trajectories or neighboring positions. Let \(L_i\) count branches assigned to path \(i\), and let \(l_{i,b}\) count branches at node \((i,b)\). CBPO updates the priority as

\begin{equation}
P_{i,b}^{\mathrm{bal}}
=
\frac{P_{i,b}^{\mathrm{raw}}}
{(1+L_i)^{\rho_{\mathrm{path}}}
 (1+l_{i,b})^{\rho_{\mathrm{node}}}},
\label{eq:balanced-priority}
\end{equation}

where \(\rho_{\mathrm{path}},\rho_{\mathrm{node}}\ge0\) control path-level and node-level decay. Before allocation, CBPO retains at most \(J_{\max}\) candidates from each parent. A candidate remains eligible while \(l_{i,b}<B_{\mathrm{node}}\) and \(L_i<B_{\mathrm{path}}\). The hard caps are necessary because power decay never reduces a priority exactly to zero. Updating both counts after each sampled branch progressively shifts allocation toward less-explored paths and nodes.

At each allocation step, CBPO selects the eligible candidate with the largest \(P^{\mathrm{bal}}\). A branch is sampled only if this score exceeds \(\kappa\). Otherwise, independent complete trajectories fill the unused rollout slots, preserving the fixed budget.

At a selected node, the trajectory is decomposed at token boundary \(b\) as

\begin{equation}
\tau=(p_b,c_b),\qquad
p_b=\tau_{<b},\qquad
c_b=\tau_{\ge b},
\label{eq:trajectory-decomposition}
\end{equation}

Branch generation holds prefix \(p_b\) fixed and resamples only continuation \(c_b\). The parent and all branches sharing \(p_b\) form an exact-prefix group \(\mathcal{G}(p_b)\). Comparing their outcomes controls for the preceding token history and isolates variation among the sampled continuations. The resulting balanced exact-prefix groups serve as input to the outcome-contrastive credit-assignment stage.

\subsection{Hierarchical Policy Optimization with CBV}\label{sec:cbv-optimization}

\begin{figure}[!t]
\centering
\includegraphics[width=0.58\linewidth]{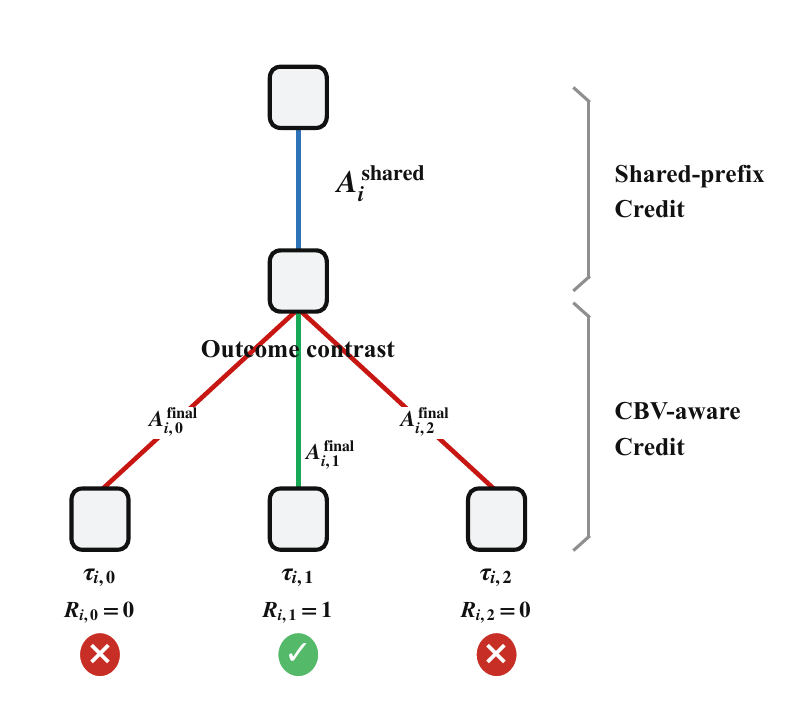}
\caption{Outcome-contrastive credit assignment. Exact-prefix continuations are compared by outcome to construct CBV-aware credit while deduplicating shared-prefix credit.}
\Description{Tree schematic in which three continuations share a prefix, receive different outcome rewards, and obtain CBV-aware continuation credit alongside deduplicated shared-prefix credit.}
\label{fig:cbv-credit-assignment}
\end{figure}

The first stage identifies positions at which the policy admits alternatives, but entropy alone cannot establish whether those alternatives affect the reward. The second stage therefore measures reward variation within each exact-prefix group and translates it into local credit. This formulation emphasizes continuations sampled at outcome-sensitive nodes while suppressing redundant gradients on their shared histories. Figure~\ref{fig:cbv-credit-assignment} illustrates the distinction between shared-prefix credit and CBV-aware continuation credit.

CBPO requires only bounded outcome rewards. Following ReTool and ToRL \cite{feng2026retool,li2025torl}, mathematical tasks use a binary correctness reward that compares trajectory answer $\hat a(\tau)$ with reference answer $a^\star$:

\begin{equation}
R(\tau)=
\begin{cases}
1, & \hat a(\tau)\text{ is symbolically equivalent to }a^\star,\\
0, & \text{otherwise}.
\end{cases}
\label{eq:binary-reward}
\end{equation}

Search tasks use normalized token-level F1 or an LLM-as-a-Judge score. Every reward satisfies $R(\tau)\in[0,1]$; Section~\ref{sec:experimental-setup} gives the benchmark-specific metrics.

\textbf{Contrastive Branch Value.} Consider exact-prefix group \(\mathcal{G}_i=\{\tau_{i,0},\ldots,\tau_{i,K_i}\}\), where \(\tau_{i,0}\) is the parent and the remaining trajectories are branches. Contrastive Branch Value (CBV) is defined as the population standard deviation of group rewards \(R_{i,0},\ldots,R_{i,K_i}\):

\begin{equation}
\operatorname{CBV}_i
=
\sqrt{
\frac{1}{K_i+1}
\sum_{k=0}^{K_i}
\left(R_{i,k}-\bar R_i\right)^2
},
\qquad
\bar R_i=\frac{1}{K_i+1}\sum_{k=0}^{K_i}R_{i,k}.
\label{eq:cbv}
\end{equation}

Larger CBV indicates greater reward variation among continuations sampled after the same prefix. Because raw CBV is nonnegative, direct addition would only increase advantage values. Standardization across valid nodes in the batch produces a comparable signed signal:

\begin{equation}
Z_i^{\mathrm{CBV}}
=
\frac{\operatorname{CBV}_i-\mu_{\mathrm{CBV}}}
{\sigma_{\mathrm{CBV}}+\epsilon}.
\label{eq:standardized-cbv}
\end{equation}

Here, \(Z_i^{\mathrm{CBV}}>0\) denotes reward variation above the batch mean. If \(\sigma_{\mathrm{CBV}}\) falls below a numerical threshold, every \(Z_i^{\mathrm{CBV}}\) is set to zero and the base advantages are retained.

\textbf{Credit assignment.}
CBV identifies relative outcome sensitivity at the group level, but the optimizer still requires token-level advantages. Problem-level GRPO normalization first yields base advantage \(A_{i,k}\) for each trajectory. The common prefix at node \(i\) receives the group-mean advantage:

\begin{equation}
A_i^{\mathrm{shared}}
=
\frac{1}{K_i+1}
\sum_{k=0}^{K_i}A_{i,k}.
\label{eq:shared-prefix-advantage}
\end{equation}

Following bounded advantage modulation \cite{he2026polarityentropy}, CBV changes only the magnitude of continuation advantage \(k\). The standardized credit is bounded as

\begin{equation}
C_{i,k}^{\mathrm{CBV}}
=
\operatorname{clip}
\left(
Z_i^{\mathrm{CBV}},
-\frac{|A_{i,k}|}{\phi},
\frac{|A_{i,k}|}{\phi}
\right).
\label{eq:bounded-cbv-credit}
\end{equation}

The CBV-aware advantage is

\begin{equation}
A_{i,k}^{\mathrm{final}}
=
A_{i,k}
+\eta\,\operatorname{sign}(A_{i,k})
\operatorname{stopgrad}
\left(C_{i,k}^{\mathrm{CBV}}\right),
\label{eq:cbv-aware-advantage}
\end{equation}

where \(\eta\) controls modulation strength and \(\phi\) limits its relative magnitude. For \(0\le\eta<\phi\), the modulated advantage retains the original sign and therefore preserves the local optimization direction.

\textbf{Hierarchical segmentation for multiple nodes.}
Let parent trajectory \(i\) contain \(J_i\) selected boundaries

\begin{equation}
b_{i,1}<b_{i,2}<\cdots<b_{i,J_i}.
\label{eq:selected-boundaries}
\end{equation}

Multiple selected ancestors can otherwise assign credit repeatedly to the same suffix. CBPO therefore partitions each parent trajectory into non-overlapping intervals, with token advantage

\begin{equation}
\widetilde A_t=
\begin{cases}
A_{i,1}^{\mathrm{shared}},
&0\le t<b_{i,1},\\[3pt]
\frac{1}{2}
\left(A_{i,j,0}^{\mathrm{final}}+A_{i,j+1}^{\mathrm{shared}}\right),
&b_{i,j}\le t<b_{i,j+1},\\[3pt]
A_{i,J_i,0}^{\mathrm{final}},
&t\ge b_{i,J_i}.
\end{cases}
\label{eq:segmented-token-advantage}
\end{equation}

where \(A_{i,j,0}^{\mathrm{final}}\) is the parent's continuation advantage at its \(j\)-th selected node. Each intermediate interval is simultaneously a suffix of the preceding node and a shared prefix of the following node. The two associated terms therefore receive equal weight.

CBPO retains the GRPO objective, replacing only trajectory-level advantage \(A_k\) with segmented token advantage \(\widetilde A_{k,t}\). For model-generated token \(y_{k,t}\), the importance ratio is
\begin{equation}
r_{k,t}(\theta)=
\frac{\pi_\theta(y_{k,t}\mid x,y_{k,<t})}
{\pi_{\mathrm{old}}(y_{k,t}\mid x,y_{k,<t})},
\label{eq:importance-ratio}
\end{equation}
and its clipped form
\begin{equation}
\bar r_{k,t}(\theta)=\operatorname{clip}
\bigl(r_{k,t}(\theta),1-\varepsilon,1+\varepsilon\bigr),
\label{eq:clipped-ratio}
\end{equation}
\begin{equation}
\ell_{k,t}(\theta)=\min\!\left(
r_{k,t}(\theta)\widetilde A_{k,t},
\bar r_{k,t}(\theta)\widetilde A_{k,t}\right).
\label{eq:clipped-surrogate}
\end{equation}
The CBPO objective is
\begin{equation}
\mathcal{J}_{\mathrm{CBPO}}(\theta)=
\mathbb{E}\!\left[
\frac{1}{M}\sum_{k=1}^{M}\frac{1}{|y_k|}\sum_{t=1}^{|y_k|}
\left(\ell_{k,t}(\theta)-\beta D_{\mathrm{KL}}
\left(\pi_\theta\,\Vert\,\pi_{\mathrm{ref}}\right)\right)
\right].
\label{eq:cbpo-objective}
\end{equation}

The objective connects the two stages: balanced exploration defines the controlled comparisons, and CBV determines their contribution to non-overlapping continuation updates. Algorithm~\ref{alg:cbpo} summarizes the end-to-end training procedure.

\begin{algorithm}[H]
\caption{Contrastive Branch Policy Optimization}
\label{alg:cbpo}
\footnotesize
\begin{algorithmic}[1]
\Input initial policy \(\pi_{\theta_{\mathrm{init}}}\), reference policy \(\pi_{\mathrm{ref}}\), data \(\mathcal D\), tools \(\mathcal T\)
\Input training steps \(S\), rollout budget \(M\), initial count \(N_0\), caps \(J_{\max},B_{\mathrm{node}},B_{\mathrm{path}}\)
\Input \(\alpha,\gamma,\kappa,\rho_{\mathrm{path}},\rho_{\mathrm{node}},d_{\min},w,K,\eta,\phi\)
\Output trained policy \(\pi_\theta\)
\State \(\pi_\theta\gets\pi_{\theta_{\mathrm{init}}}\)
\For{training step \(s=1,\ldots,S\)}
  \State \(\pi_{\mathrm{old}}\gets\pi_\theta\); sample minibatch \(\mathcal D_b\subset\mathcal D\)
  \ForAll{problems \(x\in\mathcal D_b\)}
    \State sample \(N_0\) parent trajectories; initialize rollout set \(\mathcal R_x\)
    \ForAll{parent trajectories \(\tau_i\in\mathcal R_x\)}
      \State scan \(d_{\min}\)-spaced boundaries and compute \(H_{i,b},\Delta H_{i,b},P_{i,b}^{\mathrm{raw}}\)
      \State retain the top \(J_{\max}\) candidates; set \(L_i\gets0\) and \(l_{i,b}\gets0\)
    \EndFor
    \While{\(|\mathcal R_x|<M\)}
      \State form eligible set \(\mathcal C_x=\{(i,b):L_i<B_{\mathrm{path}},\ l_{i,b}<B_{\mathrm{node}}\}\)
      \If{\(\mathcal C_x=\varnothing\)}
        \State sample \(M-|\mathcal R_x|\) independent complete trajectories; \textbf{break}
      \EndIf
      \State compute \(P_{i,b}^{\mathrm{bal}}\) and choose \((i^*,b^*)=\arg\max_{(i,b)\in\mathcal C_x}P_{i,b}^{\mathrm{bal}}\)
      \If{\(P_{i^*,b^*}^{\mathrm{bal}}>\kappa\)}
        \State fix prefix \(\tau_{i^*,<b^*}\), resample one continuation, and add it to \(\mathcal R_x\)
        \State \(L_{i^*}\gets L_{i^*}+1\); \(l_{i^*,b^*}\gets l_{i^*,b^*}+1\)
      \Else
        \State sample \(M-|\mathcal R_x|\) independent complete trajectories; \textbf{break}
      \EndIf
    \EndWhile
    \State evaluate \(R(\tau)\) and compute prompt-level base advantages \(A_{i,k}\) over \(\mathcal R_x\)
  \EndFor
  \State build exact-prefix groups; compute \(\operatorname{CBV}_i\), \(Z_i^{\mathrm{CBV}}\), \(A_i^{\mathrm{shared}}\), and \(A_{i,k}^{\mathrm{final}}\)
  \State mask each copied branch prefix; assign \(A_{i,k}^{\mathrm{final}}\) only to its sampled suffix
  \State segment each parent by Eq.~\eqref{eq:segmented-token-advantage}; keep base advantages for fallback rollouts
  \State maximize Eq.~\eqref{eq:cbpo-objective} and update \(\theta\)
\EndFor
\end{algorithmic}
\end{algorithm}

\subsection{Theoretical Analysis}\label{sec:theory}

Two properties clarify the roles of the two signals. Conditional entropy bounds the outcome information available in sampled continuations, which supports its use for candidate screening. CBV estimates conditional outcome variance under a fixed prefix, which supports its use for local credit. The bounded modulation additionally preserves the advantage sign and local PPO gradient direction.

\textbf{Property 1: generation entropy bounds attainable outcome information.}
Given exact prefix \(p\), let continuation \(C=(Y_1,\ldots,Y_L)\sim\pi_\theta(\cdot\mid p)\) produce final outcome reward \(R\). The chain rule for conditional entropy and the upper bound on conditional mutual information give

\begin{equation}
\begin{aligned}
H(C\mid p)
&=\sum_{t=1}^{L}\mathbb{E}_{Y_{<t}\mid p}
\bigl[H(Y_t\mid p,Y_{<t})\bigr],
\\
I(C;R\mid p)&\le H(C\mid p).
\end{aligned}
\label{eq:entropy-information-bound}
\end{equation}

For deterministic outcome verification, a nearly deterministic continuation can carry little information about alternative outcomes. High conditional entropy is not sufficient, however, because variation in wording, formatting, or inconsequential reasoning may leave the answer unchanged. Thus, entropy provides an upper bound rather than a direct estimate of \(I(C;R\mid p)\). CBPO uses it to screen for positions where meaningful alternatives may exist, then relies on observed branch rewards for credit assignment.

\textbf{Property 2: CBV estimates conditional outcome variance under a shared prefix.}
Suppose \(n\) continuations are sampled independently from prefix \(p\), yielding rewards \(R_1,\ldots,R_n\), and let \(\bar R=n^{-1}\sum_k R_k\). Then

\begin{equation}
\operatorname{CBV}^2(p)
=\frac{1}{n}\sum_{k=1}^{n}(R_k-\bar R)^2
=\frac{1}{2n^2}\sum_{i=1}^{n}\sum_{j=1}^{n}(R_i-R_j)^2,
\label{eq:cbv-pairwise}
\end{equation}

and

\begin{equation}
\mathbb{E}\!\left[\operatorname{CBV}^2(p)\mid p\right]
=\frac{n-1}{n}\operatorname{Var}(R\mid p).
\label{eq:cbv-expectation}
\end{equation}

The pairwise identity follows by expanding the squared differences:

\begin{equation}
\sum_{i=1}^{n}\sum_{j=1}^{n}(R_i-R_j)^2
=2n\sum_{i=1}^{n}R_i^2-2\left(\sum_{i=1}^{n}R_i\right)^2.
\label{eq:cbv-pairwise-expansion}
\end{equation}

For \(i\ne j\), conditional independence gives \(\mathbb E[(R_i-R_j)^2\mid p]=2\operatorname{Var}(R\mid p)\). The \(n\) diagonal terms are zero, leaving \(n(n-1)\) nonzero ordered pairs. Substitution into Eq.~\eqref{eq:cbv-pairwise} yields Eq.~\eqref{eq:cbv-expectation}.

Equations~\eqref{eq:cbv-pairwise} and~\eqref{eq:cbv-expectation} show that \(\tfrac{n}{n-1}\operatorname{CBV}^2(p)\) is an unbiased estimator of conditional outcome variance. Its pairwise form measures reward divergence among continuations sharing the same token history. For binary correctness with \(q_p=\Pr(R=1\mid p)\), this variance is \(q_p(1-q_p)\) and is maximal at \(q_p=1/2\). Consequently, an uncertain node with consistent outcomes receives little CBV, whereas a node separating success from failure receives more. Bounded modulation changes update magnitude within a controlled interval while preserving the advantage sign and local PPO gradient direction.

\textbf{Property 3: bounded CBV modulation preserves the update direction.}
Let \(C=C^{\mathrm{CBV}}\) satisfy the clipping constraint \(|C|\le |A|/\phi\). For any nonzero base advantage \(A\), Eq.~\eqref{eq:cbv-aware-advantage} can be written as

\begin{equation}
A^{\mathrm{final}}=wA,
\qquad
w=1+\eta\frac{C}{|A|}
\in\left[1-\frac{\eta}{\phi},1+\frac{\eta}{\phi}\right].
\label{eq:direction-preserving-weight}
\end{equation}

When \(0\le\eta<\phi\), the lower endpoint is positive. Therefore, \(A^{\mathrm{final}}\) has the same sign as \(A\). The PPO surrogate is positively homogeneous in its advantage argument, so \(\ell(r,wA)=w\ell(r,A)\) for \(w>0\). Because the modulation is stop-gradient, its local policy gradient is also scaled by the same positive \(w\). CBV consequently reallocates update magnitude without reversing the preference induced by the outcome reward. When \(A=0\), the clipping constraint forces \(C=0\), and the statement holds trivially.
\section{Experiments}\label{sec:experiments}

\subsection{Experimental Setup}\label{sec:experimental-setup}

\textbf{Datasets and evaluation.} The evaluation covers mathematical reasoning and knowledge-intensive search, two forms of tool-integrated reasoning. Each task provides access to Web Search and Python, allowing the policy to select tools without dataset-specific routing. Mathematical evaluation uses AIME 2024, AIME 2025, MATH-500, MATH \cite{hendrycks2021math}, and the complete GSM8K test set \cite{cobbe2021verifiers}. Reported metrics include Pass@1, the five-task macro average, and Pass@3/5 computed from five samples per problem. Knowledge-intensive search evaluation uses WebWalker \cite{wu2025webwalker}, HotpotQA \cite{yang2018hotpotqa}, 2WikiMultiHopQA \cite{ho2020twikimultihopqa}, MuSiQue \cite{trivedi2022musique}, and Bamboogle \cite{press2023compositionality}. WebWalker is evaluated with LLM-as-a-Judge Pass@1, whereas the other four benchmarks use normalized token-level F1. Domain-specific averages are reported because these metrics are not directly comparable.

% Main results table: sourced early so the two-column float can take the top
% of the page carrying the experimental discussion instead of being deferred
% to a float-only page at the end of the document.
\begin{table*}[!t]
\centering
\caption{Measured results (\%) for Qwen3-1.7B and Qwen3-4B under a unified protocol. Mathematical tasks report Pass@1; WebWalker reports LLM-as-a-Judge Pass@1; the other search tasks report normalized token-level F1. Math Avg. and Search Avg. are unweighted five-task means. Green subscripts show improvement over size-matched GIGPO.}
\label{tab:main-results}
\cbpomainfont
\setlength{\tabcolsep}{3.3pt}
\renewcommand{\arraystretch}{1.00}
\begin{tabular}{@{}lrrrrrrrrrrrr@{}}
\toprule[0.8pt]
\multicolumn{1}{c}{\multirow{2}{*}{\textbf{Method}}} &
\multicolumn{6}{c}{\textbf{Mathematical Reasoning}} &
\multicolumn{6}{c}{\textbf{Knowledge-Intensive Search}} \\
\cmidrule(lr){2-7}\cmidrule(lr){8-13}
& AIME24
& AIME25
& MATH500
& GSM8K
& MATH
& Math Avg.
& WebWalker
& HotpotQA
& 2Wiki
& MuSiQue
& Bamboogle
& Search Avg. \\
\midrule[0.45pt]
\multicolumn{13}{c}{\textbf{\textit{Backbone: Qwen3-1.7B}}} \\
\midrule[0.45pt]
\multicolumn{13}{l}{\rule{0pt}{2.35ex}\textbf{\textit{Training-Free Method}}} \\
Base & 13.3 & 16.7 & 82.2 & 90.8 & 86.7 & 57.9 & 5.0 & 22.0 & 25.0 & 9.5 & 34.0 & 19.1 \\
TIR Prompting & 16.7 & 23.3 & 81.4 & 88.4 & 85.1 & 59.0 & 14.5 & 36.0 & 41.0 & 16.0 & 48.0 & 31.1 \\
\cdashline{1-13}[3pt/3pt]
\multicolumn{13}{l}{\rule{0pt}{2.35ex}\textbf{\textit{Supervised and Distillation Methods}}} \\
SFT & 16.7 & 23.3 & 82.8 & 91.0 & 87.6 & 60.3 & 16.5 & 48.0 & 58.0 & 22.5 & 58.0 & 40.6 \\
OPD & 20.0 & 26.7 & 84.4 & 92.0 & 88.7 & 62.4 & 18.5 & 52.0 & 66.5 & 25.0 & 60.5 & 44.5 \\
\cdashline{1-13}[3pt/3pt]
\multicolumn{13}{l}{\rule{0pt}{2.35ex}\textbf{\textit{Classic RL Method}}} \\
GRPO & 16.7 & \underline{30.0} & 83.6 & 91.3 & 89.0 & 62.1 & 18.0 & 53.5 & 68.5 & 26.5 & 62.0 & 45.7 \\
REINFORCE++ & 20.0 & 26.7 & 83.2 & 91.9 & 88.3 & 62.0 & 18.5 & 51.0 & 65.5 & 24.0 & 61.0 & 44.0 \\
DAPO & 13.3 & 26.7 & 84.8 & 91.5 & 88.8 & 61.0 & 17.5 & 52.0 & 65.0 & 25.0 & 60.5 & 44.0 \\
GSPO & 20.0 & 26.7 & 84.6 & 91.0 & 88.7 & 62.2 & 18.5 & 54.0 & 68.0 & 26.0 & 63.0 & 45.9 \\
EAPO & \underline{23.3} & \underline{30.0} & \underline{85.6} & 91.8 & \underline{90.0} & 64.1 & 20.5 & 55.0 & 69.5 & \textbf{32.0} & 65.0 & 48.4 \\
OC-GRPO & 20.0 & \textbf{33.3} & 85.2 & \underline{92.4} & 89.5 & 64.1 & 19.5 & 54.5 & 69.0 & 27.0 & 64.0 & 46.8 \\
\cdashline{1-13}[3pt/3pt]
\multicolumn{13}{l}{\rule{0pt}{2.35ex}\textbf{\textit{Agentic RL Method}}} \\
GIGPO & \underline{23.3} & 26.7 & 84.2 & \underline{92.4} & 89.1 & $63.1_{\Delta_{\mathrm{base}}}$ & 23.5 & 58.0 & 73.0 & 29.5 & 67.0 & $50.2_{\Delta_{\mathrm{base}}}$ \\
Tree-GRPO & \underline{23.3} & \underline{30.0} & 85.0 & 92.0 & 88.8 & $63.8_{\textcolor{ForestGreen}{+1.1\%}}$ & 24.0 & 58.5 & 73.5 & 30.0 & 68.0 & $50.8_{\textcolor{ForestGreen}{+1.2\%}}$ \\
ARPO & \textbf{26.7} & \underline{30.0} & 85.0 & 91.7 & 88.3 & $\underline{64.3}_{\textcolor{ForestGreen}{+1.9\%}}$ & \underline{24.5} & \underline{59.0} & \underline{74.0} & 30.5 & \underline{69.0} & $\underline{51.4}_{\textcolor{ForestGreen}{+2.4\%}}$ \\
\rowcolor{black!10}
\textbf{CBPO} & \textbf{26.7} & \textbf{33.3} & \textbf{86.4} & \textbf{93.2} & \textbf{90.4} & $\mathbf{66.0}_{\textcolor{ForestGreen}{\mathbf{+4.6\%}}}$ & \textbf{26.0} & \textbf{61.0} & \textbf{75.5} & \underline{31.5} & \textbf{72.0} & $\mathbf{53.2}_{\textcolor{ForestGreen}{\mathbf{+6.0\%}}}$ \\
\midrule[0.45pt]
\multicolumn{13}{c}{\textbf{\textit{Backbone: Qwen3-4B}}} \\
\midrule[0.45pt]
\multicolumn{13}{l}{\rule{0pt}{2.35ex}\textbf{\textit{Training-Free Method}}} \\
Base & 13.3 & 20.0 & 84.6 & 90.0 & 89.8 & 59.5 & 7.0 & 27.0 & 31.0 & 12.0 & 40.0 & 23.4 \\
TIR Prompting & 16.7 & 26.7 & 85.1 & 92.7 & 89.3 & 62.1 & 17.5 & 41.5 & 47.0 & 19.5 & 53.5 & 35.8 \\
\cdashline{1-13}[3pt/3pt]
\multicolumn{13}{l}{\rule{0pt}{2.35ex}\textbf{\textit{Supervised and Distillation Methods}}} \\
SFT & 16.7 & 26.7 & 85.7 & 93.2 & 90.5 & 62.6 & 20.0 & 52.0 & 63.0 & 25.5 & 63.0 & 44.7 \\
OPD & 23.3 & 33.3 & 86.5 & 94.1 & 89.6 & 65.4 & 21.5 & 56.0 & 71.5 & 28.0 & 65.5 & 48.5 \\
\cdashline{1-13}[3pt/3pt]
\multicolumn{13}{l}{\rule{0pt}{2.35ex}\textbf{\textit{Classic RL Method}}} \\
GRPO & 16.7 & \textbf{40.0} & 86.0 & 93.8 & 91.3 & 65.6 & 21.5 & 57.5 & 73.0 & 29.5 & 66.5 & 49.6 \\
REINFORCE++ & \underline{26.7} & 33.3 & 85.8 & 94.2 & 91.4 & 66.3 & 22.0 & 55.0 & 70.0 & 27.0 & 65.5 & 47.9 \\
DAPO & 13.3 & 30.0 & \underline{87.8} & 93.5 & 91.5 & 63.2 & 21.0 & 56.0 & 69.0 & 28.0 & 65.0 & 47.8 \\
GSPO & 20.0 & 33.3 & 87.1 & 93.4 & 91.2 & 65.0 & 21.5 & 57.0 & 72.0 & 29.0 & 66.0 & 49.1 \\
EAPO & 23.3 & \underline{36.7} & 87.2 & 94.4 & 91.8 & 66.7 & 24.0 & 59.5 & 74.5 & \textbf{35.0} & 69.0 & 52.4 \\
OC-GRPO & 23.3 & 33.3 & 87.0 & 94.2 & 90.6 & 65.7 & 22.5 & 58.5 & 73.5 & 29.5 & 67.5 & 50.3 \\
\cdashline{1-13}[3pt/3pt]
\multicolumn{13}{l}{\rule{0pt}{2.35ex}\textbf{\textit{Agentic RL Method}}} \\
GIGPO & \underline{26.7} & 33.3 & 86.7 & \underline{94.6} & 91.0 & $66.5_{\Delta_{\mathrm{base}}}$ & 27.0 & 62.0 & 77.0 & 32.5 & 73.0 & $54.3_{\Delta_{\mathrm{base}}}$ \\
Tree-GRPO & \underline{26.7} & 33.3 & 87.5 & 94.0 & \underline{91.9} & $66.7_{\textcolor{ForestGreen}{+0.3\%}}$ & 28.0 & 62.5 & 77.5 & 33.0 & 74.0 & $55.0_{\textcolor{ForestGreen}{+1.3\%}}$ \\
ARPO & \textbf{30.0} & 33.3 & 87.4 & 94.1 & 90.7 & $\underline{67.1}_{\textcolor{ForestGreen}{+0.9\%}}$ & \underline{29.0} & \underline{63.0} & \underline{78.0} & 33.5 & \underline{75.0} & $\underline{55.7}_{\textcolor{ForestGreen}{+2.6\%}}$ \\
\rowcolor{black!10}
\textbf{CBPO} & \textbf{30.0} & \underline{36.7} & \textbf{89.5} & \textbf{96.3} & \textbf{94.0} & $\mathbf{69.3}_{\textcolor{ForestGreen}{\mathbf{+4.2\%}}}$ & \textbf{31.0} & \textbf{65.5} & \textbf{80.0} & \underline{34.5} & \textbf{76.5} & $\mathbf{57.5}_{\textcolor{ForestGreen}{\mathbf{+5.9\%}}}$ \\
\bottomrule[0.8pt]
\end{tabular}
\end{table*}

\textbf{Evaluation controls.} All methods use the same system prompt, tool interfaces, decoding settings, and answer-extraction procedure. Search calls return the top ten snippets and share an evaluation cache, preventing method-specific retrieval variation from confounding policy comparisons. Mathematical answers are checked with the same symbolic-equivalence verifier. Pass@3/5 uses five independent samples for every method and problem. AIME 2024 and AIME 2025 each contain 30 problems, so one additional correct answer changes accuracy by 3.3 percentage points. Accordingly, the analysis emphasizes cross-task averages, both model scales, and component ablations rather than small AIME differences in isolation.

\textbf{Training setup.} The experiments use Qwen3-1.7B and Qwen3-4B backbones \cite{yang2025qwen3}. Training data are drawn from Tool-Star \cite{dong2025toolstar}. Mathematical experiments use 5,000 mathematical samples, whereas search experiments use search and dual-tool trajectories from 10,000 open-domain RL samples. All reward-based methods share the SFT initialization, training data, update count, outcome rewards, tool environment, and 16 rollout slots. OPD checkpoints follow the same benchmark and decoding protocol during evaluation. CBPO allocates six slots to initial trajectories and ten to a second branching stage. Independent complete trajectories fill any unused branch slots. Candidate scanning uses \(w=20\), \(K=10\), \(d_{\min}=64\), and \(J_{\max}=3\). Allocation uses \(\alpha=0.2\), \(\gamma=2.0\), \(\kappa=0.25\), \(\rho_{\mathrm{path}}=\rho_{\mathrm{node}}=0.2\), \(B_{\mathrm{node}}=3\), and \(B_{\mathrm{path}}=4\). Credit modulation uses \(\eta=0.2\), \(\phi=2.0\), and a CBV standard-deviation threshold of \(10^{-6}\).

\textbf{Baselines.} At both model scales, the comparison includes Base, TIR prompting \cite{lin2025tir}, SFT, and on-policy distillation (OPD) \cite{li2026rethinking}. Policy-optimization baselines are GRPO \cite{shao2024deepseekmath}, REINFORCE++ \cite{hu2025reinforcepp}, DAPO \cite{yu2025dapo}, GSPO \cite{zheng2025gspo}, EAPO \cite{he2026polarityentropy}, and OC-GRPO \cite{agrawal2026ocgrpo}. Agentic and branch-based baselines are GIGPO \cite{feng2025gigpo}, Tree-GRPO \cite{ji2026treesearch}, and ARPO \cite{dong2025arpo}.

% Pass@K figure: sourced before its discussion so the wide float lands on the
% top of the following page rather than in a trailing float queue.
\begin{figure*}[!t]
\centering
\includegraphics[width=\textwidth]{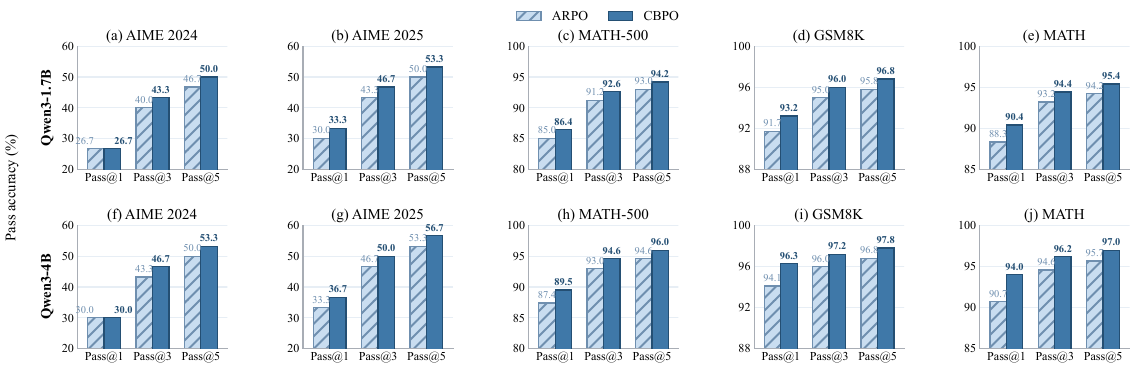}
\caption{Pass@K comparison between ARPO and CBPO. The top and bottom rows show Qwen3-1.7B and Qwen3-4B. Pass@3/5 is computed from five samples per problem. Each benchmark uses an independent vertical axis; consequently, bar heights should not be compared across benchmarks.}
\Description{Grouped bar charts comparing ARPO and CBPO at Pass@1, Pass@3, and Pass@5 on five mathematical reasoning benchmarks and two model scales.}
\label{fig:math-passk-analysis}
\end{figure*}

\subsection{Main Results}\label{sec:main-results}

The primary comparison evaluates whether the complete CBPO pipeline improves closed-form reasoning and open-world retrieval under matched rollout budgets. Table~\ref{tab:main-results} reports mathematical macro-average Pass@1 scores of 66.0\% and 69.3\% with the 1.7B and 4B backbones. These scores exceed those of ARPO, the strongest size-matched baseline by macro average, by 1.7 and 2.2 percentage points. The corresponding search averages are 53.2 and 57.5, exceeding ARPO by 1.8 points at both scales. Relative to OPD, CBPO gains 3.6 and 3.9 points on the mathematical average and 8.7 and 9.0 points on the search average. CBPO ranks first or ties for first on all five mathematical benchmarks at 1.7B and on four of five at 4B. At 4B, GRPO leads AIME 2025 by one problem. CBPO also leads four of five search benchmarks at each scale, trailing EAPO on MuSiQue by 0.5 points. This consistency across domains and model scales supports the overall pipeline. The following analyses examine the component-specific hypotheses.

\textbf{Pass@K extension.} Figure~\ref{fig:math-passk-analysis} evaluates whether the gain persists when several candidate answers are available. From five independent samples per problem, Qwen3-1.7B with CBPO reaches macro-average Pass@1, Pass@3, and Pass@5 scores of 66.0\%, 74.6\%, and 77.9\%. The corresponding Qwen3-4B scores are 69.3\%, 76.9\%, and 80.2\%. CBPO exceeds ARPO at each reported value, indicating that its advantage is not confined to single-sample decoding.

\begin{figure}[!t]
\centering
  \includegraphics[width=0.85\columnwidth]{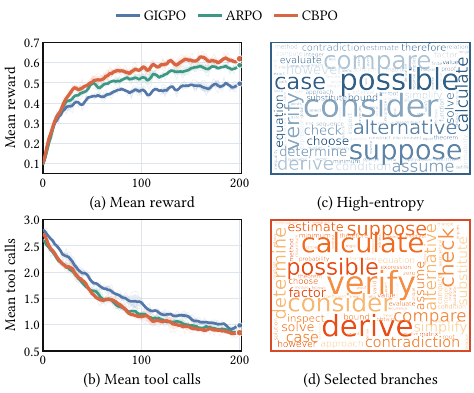}
\caption{Training dynamics and branch selection. (a) Mean reward and (b) mean tool calls per trajectory; thick lines denote smoothed means, with light traces and shading showing variation. In the qualitative word clouds, (c) token size increases monotonically with candidate entropy, whereas (d) token size increases monotonically with post-decay branching priority among retained candidates. Entropy therefore favors selection, but path- and node-level decay can change the ranking and retain some positions below the entropy-only cutoff; sizes do not encode token frequency.}
\Description{A single-column two-by-two panel grid with mean reward at upper left and mean tool calls at lower left. The upper-right qualitative word cloud sizes candidate tokens by entropy. The lower-right cloud shows retained candidates sized by post-decay branching priority, preserving an overall entropy relationship while introducing rank reversals from path- and node-level decay.}
\label{fig:training-and-branch-analysis}
\end{figure}

\textbf{Training dynamics and branch selection.} Figure~\ref{fig:training-and-branch-analysis} compares reward, tool use, and selected branch positions. CBPO attains a higher final mean reward than GIGPO and ARPO while averaging fewer tool calls. The performance gain therefore cannot be attributed to more frequent tool invocation. The qualitative examples show that decay can reorder high-entropy candidates before allocation, consistent with separating candidate discovery from budget control.

\subsection{Full-Response Candidate Analysis}\label{sec:qualitative-analysis}

Three Qwen3-1.7B trajectories from the MATH test set were inspected to locate decisions associated with their outcomes. In one successful trajectory, the model enumerated all 16 one- and two-digit candidates before calling Python. The interpreter then correctly identified the ten primes. In a failed trajectory, the submitted code also executed correctly, but the preceding reasoning listed only six parenthesizations of an arithmetic expression. Python evaluated this incomplete list and returned two values instead of the correct four. The decisive error therefore occurred in the reasoning before tool invocation, not at the tool boundary.

The third trajectory exhibited the converse pattern. Python raised the same execution exception twice, yet the model recovered the correct answer from an algebraic derivation completed before either call. A tool-return position was therefore salient but not decisive for the final reward. Together, these cases illustrate why candidate discovery should cover the full response rather than treat tool observations as universal branch boundaries. These examples provide qualitative mechanism checks, not population-level or causal evidence. A controlled tool-boundary-only ablation remains necessary to quantify this design choice independently.

% Sourced after Section 5.3 so the float lands on the next column top while
% the preceding text fills this column completely.

\subsection{Ablation Study}\label{sec:ablation}

The method motivates two quantitative predictions. Path-level and node-level decay should reduce complementary forms of allocation concentration, whereas CBV should improve learning beyond branch exploration alone. Table~\ref{tab:ablation-results} tests these predictions under matched training, rollout, and evaluation protocols.

Removing either decay level reduces the mathematical macro average by 1.5 points at both model scales. Removing both reduces it by 2.2 points, more than either individual ablation. The two decay terms therefore regulate distinct sources of allocation concentration. Removing CBV produces the largest losses, 3.0 points for Qwen3-1.7B and 2.9 points for Qwen3-4B. Branch exploration alone is therefore insufficient; branch outcomes must also inform local credit. Most of this reduction occurs on AIME 2024 and AIME 2025, the two benchmarks with the lowest absolute accuracy.

% Placed in-line below Section 5.4 when space permits; otherwise allowed to
% float so the preceding text can back-fill the previous column.
\begin{table}[!htb]
\centering
\captionsetup{font=scriptsize,skip=3pt}
\caption{Component ablations for CBPO. Values are Pass@1 (\%) and the five-task macro average. Shading marks the full method, and bold indicates the best value in each column.}
\label{tab:ablation-results}
{\scriptsize
% Wider natural columns reduce the resizebox upscaling factor, keeping the
% table full-width while lowering its rendered height.
\setlength{\tabcolsep}{5pt}
\renewcommand{\arraystretch}{1.05}
% Scale the tabular up to the full column width to remove lateral whitespace.
\resizebox{\columnwidth}{!}{%
\begin{tabular}{@{}lrrrrr>{\hspace{10pt}}r@{}}
\toprule[0.8pt]
\textbf{Method}
& AIME24
& AIME25
& MATH500
& GSM8K & MATH & Avg. \\
\midrule[0.45pt]
\multicolumn{7}{c}{\textbf{\textit{Backbone: Qwen3-1.7B}}} \\
\midrule[0.45pt]
\rowcolor{black!10}
\textbf{CBPO} & \textbf{26.7} & \textbf{33.3} & \textbf{86.4} & \textbf{93.2} & \textbf{90.4} & \textbf{66.0} \\
w/o Path-level & 23.3 & 30.0 & 86.2 & 92.8 & 90.2 & 64.5 \\
w/o Node-level & 23.3 & 30.0 & 86.1 & 93.0 & 90.3 & 64.5 \\
w/o Dual decay & 20.0 & 30.0 & 86.3 & 92.5 & 90.0 & 63.8 \\
w/o CBV & 20.0 & 26.7 & 85.5 & 92.9 & 90.1 & 63.0 \\
\midrule[0.45pt]
\multicolumn{7}{c}{\textbf{\textit{Backbone: Qwen3-4B}}} \\
\midrule[0.45pt]
\rowcolor{black!10}
\textbf{CBPO} & \textbf{30.0} & \textbf{36.7} & \textbf{89.5} & \textbf{96.3} & \textbf{94.0} & \textbf{69.3} \\
w/o Path-level & 26.7 & 33.3 & 89.3 & 95.9 & 93.8 & 67.8 \\
w/o Node-level & 26.7 & 33.3 & 89.0 & 96.0 & 93.9 & 67.8 \\
w/o Dual decay & 23.3 & 33.3 & 89.4 & 95.7 & 93.9 & 67.1 \\
w/o CBV & 23.3 & 30.0 & 89.0 & 96.0 & 93.7 & 66.4 \\
\bottomrule[0.8pt]
\end{tabular}}%
}
\end{table}

% Stacked directly below Table 2 so both tables fill the column before the
% Branching Configuration discussion begins.
\begin{table}[!htb]
\centering
\captionsetup{font=scriptsize,skip=3pt}
\caption{CBPO Pass@1 (\%) under different rollout configurations. \(M=N+B\). Shading marks the default configuration, and bold indicates the best value in each column.}
\label{tab:branching-configuration}
{\scriptsize
% Wider natural columns reduce the resizebox upscaling factor, keeping the
% table full-width while lowering its rendered height.
\setlength{\tabcolsep}{3pt}
\renewcommand{\arraystretch}{1.00}
% Scale the tabular up to the full column width to remove lateral whitespace.
\resizebox{\columnwidth}{!}{%
\begin{tabular}{@{}ccc*{5}{r}>{\hspace{9pt}}r@{}}
\toprule[0.8pt]
\multicolumn{3}{c}{\textbf{Rollout allocation}} &
\multicolumn{6}{c}{\textbf{Pass@1 accuracy (\%)}} \\
\cmidrule(lr){1-3}\cmidrule(lr){4-9}
\shortstack{Total\\$M$} & \shortstack{Initial\\$N$} & \shortstack{Branch\\$B$} &
AIME24 & AIME25 & MATH500 & GSM8K & MATH & Avg. \\
\midrule[0.45pt]
\multicolumn{9}{c}{\textbf{\textit{Backbone: Qwen3-1.7B}}} \\
\midrule[0.45pt]
4 & 1 & 3 & 13.3 & 23.3 & 82.2 & 89.9 & 86.0 & 58.9 \\
4 & 2 & 2 & 16.7 & 23.3 & 82.8 & 90.5 & 86.8 & 60.0 \\
\cdashline{1-9}[3pt/3pt]
8 & 2 & 6 & 16.7 & 26.7 & 83.7 & 91.0 & 87.8 & 61.2 \\
8 & 4 & 4 & 20.0 & 26.7 & 84.0 & 91.5 & 88.3 & 62.1 \\
\cdashline{1-9}[3pt/3pt]
16 & 4 & 12 & 23.3 & \textbf{36.7} & 85.8 & 92.9 & 89.8 & 65.7 \\
\rowcolor{black!10}
\textbf{16} & \textbf{6} & \textbf{10} & \textbf{26.7} & 33.3 & \textbf{86.4} & \textbf{93.2} & \textbf{90.4} & \textbf{66.0} \\
16 & 8 & 8 & \textbf{26.7} & 30.0 & 86.1 & 93.0 & 90.2 & 65.2 \\
16 & 12 & 4 & 23.3 & 30.0 & 85.8 & 92.6 & 89.8 & 64.3 \\
\midrule[0.45pt]
\multicolumn{9}{c}{\textbf{\textit{Backbone: Qwen3-4B}}} \\
\midrule[0.45pt]
4 & 1 & 3 & 16.7 & 26.7 & 84.4 & 91.8 & 89.0 & 61.7 \\
4 & 2 & 2 & 20.0 & 26.7 & 85.0 & 92.3 & 89.7 & 62.7 \\
\cdashline{1-9}[3pt/3pt]
8 & 2 & 6 & 20.0 & 30.0 & 86.6 & 93.8 & 91.0 & 64.3 \\
8 & 4 & 4 & 23.3 & 30.0 & 87.0 & 94.2 & 91.5 & 65.2 \\
\cdashline{1-9}[3pt/3pt]
16 & 4 & 12 & 26.7 & \textbf{40.0} & 88.9 & 95.8 & 93.3 & 68.9 \\
\rowcolor{black!10}
\textbf{16} & \textbf{6} & \textbf{10} & \textbf{30.0} & 36.7 & \textbf{89.5} & \textbf{96.3} & 94.0 & \textbf{69.3} \\
16 & 8 & 8 & \textbf{30.0} & 33.3 & 89.2 & 96.0 & \textbf{94.4} & 68.6 \\
16 & 12 & 4 & 26.7 & 33.3 & 88.8 & 95.6 & 93.2 & 67.5 \\
\bottomrule[0.8pt]
\end{tabular}}%
}
\end{table}

\subsection{Branching Configuration}\label{sec:branching-configuration}

Balanced branching requires sufficient independent paths for global coverage and sufficient shared-prefix continuations for local comparison. This trade-off is evaluated by varying the allocation between initial trajectories \(N\) and dynamic branches \(B\). Total budgets are \(M=N+B\in\{4,8,16\}\), with all other settings held fixed.

Increasing \(M\) from 4 to 16 improves the macro average at both model scales. Within a fixed budget, too few initial trajectories restrict path coverage, whereas too few branches weaken exact-prefix comparisons. The highest macro average occurs at \(N=6,B=10\) for both backbones. This shared optimum favors branch sampling while retaining sufficient independent trajectories for global coverage.

% Finish the experiment float queue before the closing material.  This keeps
% the conclusion and bibliography contiguous instead of allowing large result
% floats to split the references across several visually sparse pages.
\FloatBarrier
\section{Conclusion}\label{sec:conclusion}

CBPO addresses fine-grained credit assignment by separating two operations that branch-based RL often conflates. Full-response entropy scanning with path-level and node-level decay allocates a fixed rollout budget. Reward variation within exact-prefix groups then modulates the contribution of each sampled continuation. Prefix masking and hierarchical segmentation prevent shared tokens from receiving duplicated credit. With Qwen3-1.7B and Qwen3-4B, CBPO attains mathematical macro averages of 66.0\% and 69.3\% and search averages of 53.2 and 57.5. Cross-scale ablations attribute distinct gains to balanced allocation and CBV. However, the independent contribution of full-response scanning still requires a controlled boundary-only ablation. Within the evaluated models, tasks, and budgets, the evidence supports a bounded design principle: uncertainty identifies alternatives, whereas observed outcome variation provides a stronger signal for local credit assignment.

% Let deferred experiment floats share the remaining pages with the declaration
% and references; their order is still preserved by LaTeX's float queues.
\section*{Ethical Considerations}

{\small
The study uses public benchmarks and collects no new personal or human-subject data. Open-web retrieval may nevertheless expose a model to inaccurate, biased, offensive, or privacy-sensitive material. Benchmark accuracy does not establish the reliability or social neutrality of retrieved sources. Generated Python code runs in a sandbox, and tool calls are capped to limit security and resource risks. Like other branch-based methods, CBPO samples several continuations and therefore incurs additional generation cost. A fixed rollout budget bounds this cost, and all methods are compared using matched rollout slots. The experiments evaluate benchmark accuracy rather than deployment safety. High-stakes applications would require domain-specific testing, content filtering, access controls, and human oversight.\par
}

\renewcommand{\bibfont}{\fontsize{5.5pt}{5.8pt}\selectfont}
\setlength{\bibsep}{0pt plus 0.1pt}
\bibliographystyle{ACM-Reference-Format}
\bibliography{cbpo_references}

@inproceedings{wei2022chain,
  author =        {J. Wei and others},
  booktitle =     {Advances in Neural Information Processing Systems},
  title =         {{Chain-of-Thought Prompting Elicits Reasoning in
                   Large Language Models}},
  year =          {2022},
}

@misc{schulman2017ppo,
  author =        {J. Schulman and others},
  howpublished =  {arXiv preprint arXiv:1707.06347},
  title =         {{Proximal Policy Optimization Algorithms}},
  year =          {2017},
  doi =           {10.48550/arXiv.1707.06347},
  url =           {https://arxiv.org/abs/1707.06347},
}

@misc{shao2024deepseekmath,
  author =        {Z. Shao and others},
  howpublished =  {arXiv preprint arXiv:2402.03300},
  title =         {{DeepSeekMath: Pushing the Limits of Mathematical
                   Reasoning in Open Language Models}},
  year =          {2024},
  doi =           {10.48550/arXiv.2402.03300},
  url =           {https://arxiv.org/abs/2402.03300},
}

@misc{guo2025deepseekr1,
  author =        {D. Guo and others},
  howpublished =  {arXiv preprint arXiv:2501.12948},
  title =         {{DeepSeek-R1: Incentivizing Reasoning Capability in
                   LLMs via Reinforcement Learning}},
  year =          {2025},
  doi =           {10.48550/arXiv.2501.12948},
  url =           {https://arxiv.org/abs/2501.12948},
}

@misc{yu2025dapo,
  author =        {Q. Yu and others},
  howpublished =  {arXiv preprint arXiv:2503.14476},
  title =         {{DAPO: An Open-Source LLM Reinforcement Learning
                   System at Scale}},
  year =          {2025},
  doi =           {10.48550/arXiv.2503.14476},
  url =           {https://arxiv.org/abs/2503.14476},
}

@misc{hu2025reinforcepp,
  author =        {J. Hu and others},
  howpublished =  {arXiv preprint arXiv:2501.03262},
  title =         {{REINFORCE++: Stabilizing Critic-Free Policy
                   Optimization with Global Advantage Normalization}},
  year =          {2025},
  doi =           {10.48550/arXiv.2501.03262},
  url =           {https://arxiv.org/abs/2501.03262},
}

@misc{li2026rethinking,
  author =        {Y. Li and others},
  howpublished =  {arXiv preprint arXiv:2604.13016},
  title =         {{Rethinking On-Policy Distillation of Large Language
                   Models: Phenomenology, Mechanism, and Recipe}},
  year =          {2026},
  doi =           {10.48550/arXiv.2604.13016},
  url =           {https://arxiv.org/abs/2604.13016},
}

@inproceedings{yao2023react,
  author =        {S. Yao and others},
  booktitle =     {International Conference on Learning Representations},
  title =         {{ReAct: Synergizing Reasoning and Acting in Language
                   Models}},
  year =          {2023},
}

@inproceedings{gou2024tora,
  author =        {Z. Gou and others},
  booktitle =     {International Conference on Learning Representations},
  title =         {{ToRA: A Tool-Integrated Reasoning Agent for
                   Mathematical Problem Solving}},
  year =          {2024},
}

@inproceedings{feng2026retool,
  author =        {J. Feng and others},
  booktitle =     {International Conference on Learning Representations},
  title =         {{ReTool: Reinforcement Learning for Strategic Tool
                   Use in LLMs}},
  year =          {2026},
}

@misc{li2025torl,
  author =        {X. Li and H. Zou and P. Liu},
  howpublished =  {arXiv preprint arXiv:2503.23383},
  title =         {{ToRL: Scaling Tool-Integrated Reinforcement
                   Learning}},
  year =          {2025},
  doi =           {10.48550/arXiv.2503.23383},
  url =           {https://arxiv.org/abs/2503.23383},
}

@misc{dong2025toolstar,
  author =        {G. Dong and others},
  howpublished =  {arXiv preprint arXiv:2505.16410},
  title =         {{Tool-Star: Empowering LLM-Brained Multi-Tool
                   Reasoner via Reinforcement Learning}},
  year =          {2025},
  doi =           {10.48550/arXiv.2505.16410},
  url =           {https://arxiv.org/abs/2505.16410},
}

@inproceedings{lightman2024verify,
  author =        {H. Lightman and others},
  booktitle =     {International Conference on Learning Representations},
  title =         {{Let's Verify Step by Step}},
  year =          {2024},
}

@misc{lai2024stepdpo,
  author =        {X. Lai and others},
  howpublished =  {arXiv preprint arXiv:2406.18629},
  title =         {{Step-DPO: Step-Wise Preference Optimization for
                   Long-Chain Reasoning of LLMs}},
  year =          {2024},
  doi =           {10.48550/arXiv.2406.18629},
  url =           {https://arxiv.org/abs/2406.18629},
}

@inproceedings{hou2025treerl,
  author =        {Z. Hou and Z. Hu and Y. Li and R. Lu and J. Tang and
                   Y. Dong},
  booktitle =     {Proceedings of the Annual Meeting of the Association
                   for Computational Linguistics},
  title =         {{TreeRL: LLM Reinforcement Learning with On-Policy
                   Tree Search}},
  year =          {2025},
}

@misc{feng2025gigpo,
  author =        {L. Feng and Z. Xue and T. Liu and B. An},
  howpublished =  {arXiv preprint arXiv:2505.10978},
  title =         {{Group-in-Group Policy Optimization for LLM Agent
                   Training}},
  year =          {2025},
  doi =           {10.48550/arXiv.2505.10978},
  url =           {https://arxiv.org/abs/2505.10978},
}

@inproceedings{ji2026treesearch,
  author =        {Y. Ji and Z. Ma and Y. Wang and G. Chen and X. Chu and
                   L. Wu},
  booktitle =     {International Conference on Learning Representations},
  title =         {{Tree Search for LLM Agent Reinforcement Learning}},
  year =          {2026},
}

@misc{dong2025arpo,
  author =        {G. Dong and others},
  howpublished =  {arXiv preprint arXiv:2507.19849},
  title =         {{Agentic Reinforced Policy Optimization}},
  year =          {2025},
  doi =           {10.48550/arXiv.2507.19849},
  url =           {https://arxiv.org/abs/2507.19849},
}

@misc{lin2024critical,
  author =        {Z. Lin and T. Liang and J. Xu and X. Wang and R. Luo and
                   C. Shi and S. Li and Y. Yang and Z. Tu},
  howpublished =  {arXiv preprint arXiv:2411.19943},
  title =         {{Critical Tokens Matter: Token-Level Contrastive
                   Estimation Enhances LLM's Reasoning Capability}},
  year =          {2024},
  doi =           {10.48550/arXiv.2411.19943},
  url =           {https://arxiv.org/abs/2411.19943},
}

@misc{cheng2025exploration,
  author =        {D. Cheng and S. Huang and X. Zhu and B. Dai and
                   W.~X. Zhao and Z. Zhang and F. Wei},
  howpublished =  {arXiv preprint arXiv:2506.14758},
  title =         {{Reasoning with Exploration: An Entropy Perspective}},
  year =          {2025},
  doi =           {10.48550/arXiv.2506.14758},
  url =           {https://arxiv.org/abs/2506.14758},
}

@misc{wang2025entropyminority,
  author =        {S. Wang and others},
  howpublished =  {arXiv preprint arXiv:2506.01939},
  title =         {{Beyond the 80/20 Rule: High-Entropy Minority Tokens
                   Drive Effective Reinforcement Learning for LLM
                   Reasoning}},
  year =          {2025},
  doi =           {10.48550/arXiv.2506.01939},
  url =           {https://arxiv.org/abs/2506.01939},
}

@misc{cobbe2021verifiers,
  author =        {K. Cobbe and others},
  howpublished =  {arXiv preprint arXiv:2110.14168},
  title =         {{Training Verifiers to Solve Math Word Problems}},
  year =          {2021},
  doi =           {10.48550/arXiv.2110.14168},
  url =           {https://arxiv.org/abs/2110.14168},
}

@misc{hendrycks2021math,
  author =        {D. Hendrycks and others},
  howpublished =  {arXiv preprint arXiv:2103.03874},
  title =         {{Measuring Mathematical Problem Solving with the MATH
                   Dataset}},
  year =          {2021},
  doi =           {10.48550/arXiv.2103.03874},
  url =           {https://arxiv.org/abs/2103.03874},
}

@inproceedings{wu2025webwalker,
  address =       {Vienna, Austria},
  author =        {Wu, Jialong and Yin, Wenbiao and Jiang, Yong and
                   Wang, Zhenglin and Xi, Zekun and Fang, Runnan and
                   Zhang, Linhai and He, Yulan and Zhou, Deyu and
                   Xie, Pengjun and Huang, Fei},
  booktitle =     {Proceedings of the 63rd Annual Meeting of the
                   Association for Computational Linguistics (Volume 1:
                   Long Papers)},
  pages =         {10290--10305},
  publisher =     {Association for Computational Linguistics},
  title =         {{{WebWalker}: Benchmarking {LLM}s in Web Traversal}},
  year =          {2025},
  doi =           {10.18653/v1/2025.acl-long.508},
  url =           {https://aclanthology.org/2025.acl-long.508/},
}

@inproceedings{yang2018hotpotqa,
  address =       {Brussels, Belgium},
  author =        {Yang, Zhilin and Qi, Peng and Zhang, Saizheng and
                   Bengio, Yoshua and Cohen, William W. and
                   Salakhutdinov, Ruslan and Manning, Christopher D.},
  booktitle =     {Proceedings of the 2018 Conference on Empirical
                   Methods in Natural Language Processing},
  pages =         {2369--2380},
  publisher =     {Association for Computational Linguistics},
  title =         {{{HotpotQA}: A Dataset for Diverse, Explainable
                   Multi-hop Question Answering}},
  year =          {2018},
  doi =           {10.18653/v1/D18-1259},
  url =           {https://aclanthology.org/D18-1259/},
}

@inproceedings{ho2020twikimultihopqa,
  address =       {Barcelona, Spain (Online)},
  author =        {Ho, Xanh and Duong Nguyen, Anh-Khoa and
                   Sugawara, Saku and Aizawa, Akiko},
  booktitle =     {Proceedings of the 28th International Conference on
                   Computational Linguistics},
  pages =         {6609--6625},
  publisher =     {International Committee on Computational Linguistics},
  title =         {{Constructing A Multi-hop {QA} Dataset for
                   Comprehensive Evaluation of Reasoning Steps}},
  year =          {2020},
  doi =           {10.18653/v1/2020.coling-main.580},
  url =           {https://aclanthology.org/2020.coling-main.580/},
}

@article{trivedi2022musique,
  author =        {Trivedi, Harsh and Balasubramanian, Niranjan and
                   Khot, Tushar and Sabharwal, Ashish},
  journal =       {Transactions of the Association for Computational
                   Linguistics},
  pages =         {539--554},
  title =         {{{MuSiQue}: Multihop Questions via Single-hop
                   Question Composition}},
  volume =        {10},
  year =          {2022},
  doi =           {10.1162/tacl_a_00475},
  url =           {https://aclanthology.org/2022.tacl-1.31/},
}

@inproceedings{press2023compositionality,
  address =       {Singapore},
  author =        {Press, Ofir and Zhang, Muru and Min, Sewon and
                   Schmidt, Ludwig and Smith, Noah A. and Lewis, Mike},
  booktitle =     {Findings of the Association for Computational
                   Linguistics: EMNLP 2023},
  pages =         {5687--5711},
  publisher =     {Association for Computational Linguistics},
  title =         {{Measuring and Narrowing the Compositionality Gap in
                   Language Models}},
  year =          {2023},
  doi =           {10.18653/v1/2023.findings-emnlp.378},
  url =           {https://aclanthology.org/2023.findings-emnlp.378/},
}

@misc{zheng2025gspo,
  author =        {C. Zheng and S. Liu and M. Li and X.-H. Chen and
                   B. Yu and C. Gao and K. Dang and Y. Liu and R. Men and
                   A. Yang and J. Zhou and J. Lin},
  howpublished =  {arXiv preprint arXiv:2507.18071},
  title =         {{Group Sequence Policy Optimization}},
  year =          {2025},
  doi =           {10.48550/arXiv.2507.18071},
  url =           {https://arxiv.org/abs/2507.18071},
}

@inproceedings{rafailov2023dpo,
  author =        {R. Rafailov and A. Sharma and E. Mitchell and
                   C.~D. Manning and S. Ermon and C. Finn},
  booktitle =     {Advances in Neural Information Processing Systems},
  title =         {{Direct Preference Optimization: Your Language Model
                   Is Secretly a Reward Model}},
  year =          {2023},
}

@misc{he2026polarityentropy,
  author =        {Y. He and H. Wu and S. Liu and H. Ge and H. Zhou and
                   K. Wu and Z. Zheng and Q. Lin and Z. Zhong and
                   Y. Zhang},
  howpublished =  {arXiv preprint arXiv:2604.11056},
  title =         {{Rethinking Token-Level Credit Assignment in RLVR: A
                   Polarity-Entropy Analysis}},
  year =          {2026},
  doi =           {10.48550/arXiv.2604.11056},
  url =           {https://arxiv.org/abs/2604.11056},
}

@misc{dong2025aepo,
  author =        {G. Dong and L. Bao and Z. Wang and K. Zhao and X. Li and
                   J. Jin and J. Yang and H. Mao and F. Zhang and K. Gai and
                   G. Zhou and Y. Zhu and J.-R. Wen and Z. Dou},
  howpublished =  {arXiv preprint arXiv:2510.14545},
  title =         {{Agentic Entropy-Balanced Policy Optimization}},
  year =          {2025},
  doi =           {10.48550/arXiv.2510.14545},
  url =           {https://arxiv.org/abs/2510.14545},
}

@misc{lin2025tir,
  author =        {H. Lin and Z. Xu},
  howpublished =  {arXiv preprint arXiv:2508.19201},
  title =         {{Understanding Tool-Integrated Reasoning}},
  year =          {2025},
  doi =           {10.48550/arXiv.2508.19201},
  url =           {https://arxiv.org/abs/2508.19201},
}

@inproceedings{jin2025searchr1,
  author =        {B. Jin and H. Zeng and Z. Yue and J. Yoon and
                   S.~O. Arik and D. Wang and H. Zamani and J. Han},
  booktitle =     {Conference on Language Modeling},
  title =         {{Search-R1: Training LLMs to Reason and Leverage
                   Search Engines with Reinforcement Learning}},
  year =          {2025},
}

@misc{sun2025zerosearch,
  author =        {H. Sun and Z. Qiao and J. Guo and X. Fan and Y. Hou and
                   Y. Jiang and P. Xie and Y. Zhang and F. Huang and
                   J. Zhou},
  howpublished =  {arXiv preprint arXiv:2505.04588},
  title =         {{ZeroSearch: Incentivize the Search Capability of
                   LLMs without Searching}},
  year =          {2025},
  doi =           {10.48550/arXiv.2505.04588},
  url =           {https://arxiv.org/abs/2505.04588},
}

@misc{yang2025qwen3,
  author =        {A. Yang and others},
  howpublished =  {arXiv preprint arXiv:2505.09388},
  title =         {{Qwen3 Technical Report}},
  year =          {2025},
  doi =           {10.48550/arXiv.2505.09388},
  url =           {https://arxiv.org/abs/2505.09388},
}

@misc{agrawal2026ocgrpo,
  author =        {P. Agrawal and A. Samanta and S. Ghasemlou and
                   B. Vidolov and J. Bhandari and K. Asadi and D. Jiang and
                   A. Modi},
  howpublished =  {arXiv preprint arXiv:2607.19313},
  title =         {{Off-Context GRPO: Learning to Reason on Hard
                   Problems using Privileged Information}},
  year =          {2026},
  doi =           {10.48550/arXiv.2607.19313},
  url =           {https://arxiv.org/abs/2607.19313},
}

@inproceedings{jiang2026tablemind,
  address =       {Boise, ID, USA},
  author =        {Jiang, Chuang and Cheng, Mingyue and Tao, Xiaoyu and
                   Mao, Qingyang and Ouyang, Jie and Liu, Qi},
  booktitle =     {Proceedings of the Nineteenth ACM International
                   Conference on Web Search and Data Mining},
  pages =         {260--270},
  publisher =     {ACM},
  title =         {{TableMind: An Autonomous Programmatic Agent for
                   Tool-Augmented Table Reasoning}},
  year =          {2026},
  doi =           {10.1145/3773966.3777932},
  url =           {https://doi.org/10.1145/3773966.3777932},
}

@inproceedings{zhang2026toolcure,
  address =       {Boise, ID, USA},
  author =        {Zhang, Jie and Bi, Dongsheng and Sun, Tao and
                   Yang, Minghui and Wang, Jian and Wang, Yiwei},
  booktitle =     {Proceedings of the Nineteenth ACM International
                   Conference on Web Search and Data Mining},
  pages =         {946--954},
  publisher =     {ACM},
  title =         {{TOOL-CURE: Tool Selection via Curriculum-Enhanced
                   Reinforcement Learning with Sample Screening for
                   LLMs}},
  year =          {2026},
  doi =           {10.1145/3773966.3777952},
  url =           {https://doi.org/10.1145/3773966.3777952},
}

@inproceedings{zhang2026knowfc,
  address =       {Boise, ID, USA},
  author =        {Zhang, Yue and Zhou, Shicheng and Li, Xiaopeng and
                   Tian, Zhiliang and Gao, Yifu and Zhang, Shiyi and
                   Hou, Wenqing and Liu, Yuying and Zhou, Bin},
  booktitle =     {Proceedings of the Nineteenth ACM International
                   Conference on Web Search and Data Mining},
  pages =         {996--1006},
  publisher =     {ACM},
  title =         {{KnowFC: Navigating Knowledge Conflicts in Large
                   Language Model-based Fact-Checking}},
  year =          {2026},
  doi =           {10.1145/3773966.3777935},
  url =           {https://doi.org/10.1145/3773966.3777935},
}

@inproceedings{rathee2025quam,
  address =       {Hannover, Germany},
  author =        {Rathee, Mandeep and MacAvaney, Sean and Anand, Avishek},
  booktitle =     {Proceedings of the Eighteenth ACM International
                   Conference on Web Search and Data Mining},
  pages =         {954--962},
  publisher =     {ACM},
  title =         {{Quam: Adaptive Retrieval through Query Affinity
                   Modelling}},
  year =          {2025},
  doi =           {10.1145/3701551.3703584},
  url =           {https://doi.org/10.1145/3701551.3703584},
}

\end{document}